\documentclass[conference]{IEEEtran}
\IEEEoverridecommandlockouts
\usepackage{cite}
\usepackage{amsmath,amssymb,amsfonts}
\usepackage{algorithmic}
\usepackage{graphicx}
\usepackage{textcomp}
\usepackage{xcolor}
\graphicspath{{..figures/}{../figures/}}
\DeclareGraphicsExtensions{.pdf,.jpeg,.png}
\usepackage{siunitx}
\def\BibTeX{{\rm B\kern-.05em{\sc i\kern-.025em b}\kern-.08em
    T\kern-.1667em\lower.7ex\hbox{E}\kern-.125emX}}

\newcommand{\blockcomment}[1]{}

\begin{document}

\title{Semantic Landmark Detection \& Classification Using Neural Networks For 3D In-Air Sonar}

\blockcomment{
\author{\IEEEauthorblockN{Wouter Jansen}
\IEEEauthorblockA{\textit{FTI Cosys-Lab, University of Antwerp}\\ Antwerp, Belgium \\
\textit{Flanders Make Strategic Research Centre}\\ Lommel, Belgium\\
wouter.jansen@uantwerpen.be}
\and
\IEEEauthorblockN{Jan Steckel}
\IEEEauthorblockA{\textit{FTI Cosys-Lab, University of Antwerp}\\ Antwerp, Belgium \\
\textit{Flanders Make Strategic Research Centre}\\ Lommel, Belgium\\
jan.steckel@uantwerpen.be}
}
}

\author{
    \IEEEauthorblockN{
        Wouter Jansen\IEEEauthorrefmark{1}\IEEEauthorrefmark{2}\IEEEauthorrefmark{3}
        Jan Steckel\IEEEauthorrefmark{1}\IEEEauthorrefmark{2}\IEEEauthorrefmark{4}
    }
    \IEEEauthorblockA{\IEEEauthorrefmark{1}Cosys-Lab, Faculty of Applied Engineering, University of Antwerp, Antwerp, Belgium}
    \IEEEauthorblockA{\IEEEauthorrefmark{2}Flanders Make Strategic Research Centre, Lommel, Belgium}
    \IEEEauthorblockA{\IEEEauthorrefmark{3}wouter.jansen@uantwerpen.be}
    \IEEEauthorblockA{\IEEEauthorrefmark{4}jan.steckel@uantwerpen.be}
 }

\maketitle

\begin{abstract}
In challenging environments where traditional sensing modalities struggle, in-air sonar offers resilience to optical interference. Placing a priori known landmarks in these environments can eliminate accumulated errors in autonomous mobile systems such as Simultaneous Localization and Mapping (SLAM) and autonomous navigation. We present a novel approach using a convolutional neural network to detect and classify ten different reflector landmarks with varying radii using in-air 3D sonar. Additionally, the network predicts the orientation angle of the detected landmarks. The neural network is trained on cochleograms, representing echoes received by the sensor in a time-frequency domain. Experimental results in cluttered indoor settings show promising performance. The CNN achieves a 97.3\% classification accuracy on the test dataset, accurately detecting both the presence and absence of landmarks. Moreover, the network predicts landmark orientation angles with an RMSE lower than \SI{10}{\degree}, enhancing the utility in SLAM and autonomous navigation applications. This advancement improves the robustness and accuracy of autonomous systems in challenging environments.
\end{abstract}


\section{Introduction}\label{sec:intro}

In the field of mobile autonomous systems for sectors such as manufacturing, agriculture, port operations, search \& rescue, mining and other heavy industries, challenging conditions continue to be an obstacle for light-based sensing modalities such as LiDAR and cameras \cite{yoneda_automated_2019, zhang_perception_2023, mavridou_machine_2019, thombre_sensors_2022, debeunne_review_2020, bac_harvesting_2014}. In harsh environmental conditions (fog, dust, rain, snow, dirt), other sensing modalities, such as in-air sonar, can continue functioning unaffected by optical or electromagnetic interference. 
\\Ultrasonic sensing, known for its low-cost distance estimation in its more simple form, has been further developed to be capable of 3D high angular resolution scanning with dense microphone arrays \cite{laurijssen_hiris_2024, kerstens_ertis_2019, allevato_air-coupled_2022}. This advancement facilitates detecting the environment in greater detail and differentiating nearby objects. Furthermore, it enables mobile robotics applications such as Simultaneous Localization and Mapping (SLAM) and autonomous navigation \cite{jansen_real-time_2022, steckel_biomimetic_2012, everett_sensors_1995}. Recognizing the type of and being able to self-localize based on specific landmarks can significantly enhance these applications \cite{thrun_probabilistic_2005} and can correct the accumulated error in SLAM \cite{grisetti_tutorial_2010}. For ultrasonic sensing especially, where the modality is vulnerable to ambiguous measurements. However, semantically understanding the observed environment is a significant challenge with ultrasonic sensing. Ultrasonic echoes vary based on object structure, surface material, and orientation \cite{werner_g_neubauer_acoustic_1986, simmons_acoustic_1989}. These variations affect the echo's amplitude, time structure, spectrum, and phase, influencing object classification \cite{eisele_relevance_2024}. An example of these echo variations can be seen in echolocating bats that use their emitted broadband signals to identify objects in lushly vegetated environments\cite{von_helversen_object_2004}. 
\\In recent years, research into ultrasonic semantic perception has found methods for place recognition environments \cite{vanderelst_place_2016, barat_classification_2001}, acoustic source classification \cite{zaheer_survey_2023, sabatini_digital-signal-processing_2001, schenck_airleakslam_2019}, and object surface analysis \cite{bouhamed_possibilistic_2014, zhu_surrounding_2023}. Multi-object classification has been heavily explored using neural networks and works in distinguishing landmarks with ultrasonic sensors \cite{ kroh_classification_2019, dror_three-dimensional_1995, ayrulu_neural_2001, kroh_evaluation_2020}. The object classes used are often those with shapes that exhibit noticeable differences in their sound reflection behavior, such as (acute) corners, flat surfaces, sharp edges, and spheres. However, this limits the information you can encode in different landmarks. In comparison, in the optical spectrum, a popular landmark used is QR-codes, which can encode 2953 bytes of information. 
\\A potential solution would be to use shapes that can vary in size, and that can still be uniquely classified with acoustic sensors. One such shape also seen in nature is the dish-shaped leaf of the Cuban liana \textit{Marcgravia evenia}. This bat-pollinated plant can lure echolocating bats because of its conspicuous acoustic echo. In previous research, these leaves were approximated using 3D-printed shapes with various radius, depth, and edge finishes and could be localized using a multi-transducer array \cite{simon_bioinspired_2020}. A support vector machine (SVM) was used for a more real-time application for landmark classification and detection with four different radius variations, reaching results of around 67.2\% accuracy \cite{de_backer_detecting_2023}.\\ In this paper, we expand upon the work of \cite{de_backer_detecting_2023} and use a convolution neural network with supervised learning on a large dataset to detect and classify between ten different reflectors, each exhibiting a unique radius. Notably, the radii values are closely clustered, with a maximum difference of only \SI{5}{\mm} between any two shapes. This neural network uses a bio-inspired representation of the reflected echoes in the form of a cochleogram. This time-frequency representation is based on modeling the cochlea of biological species such as an echolocating bat \cite{valero_gammatone_2012, patterson_r_d_efficient_1987}. The following sections will detail our methodology, experimental results, and conclusions. 

\begin{figure}
    \centering
    \includegraphics[width=0.95\linewidth]{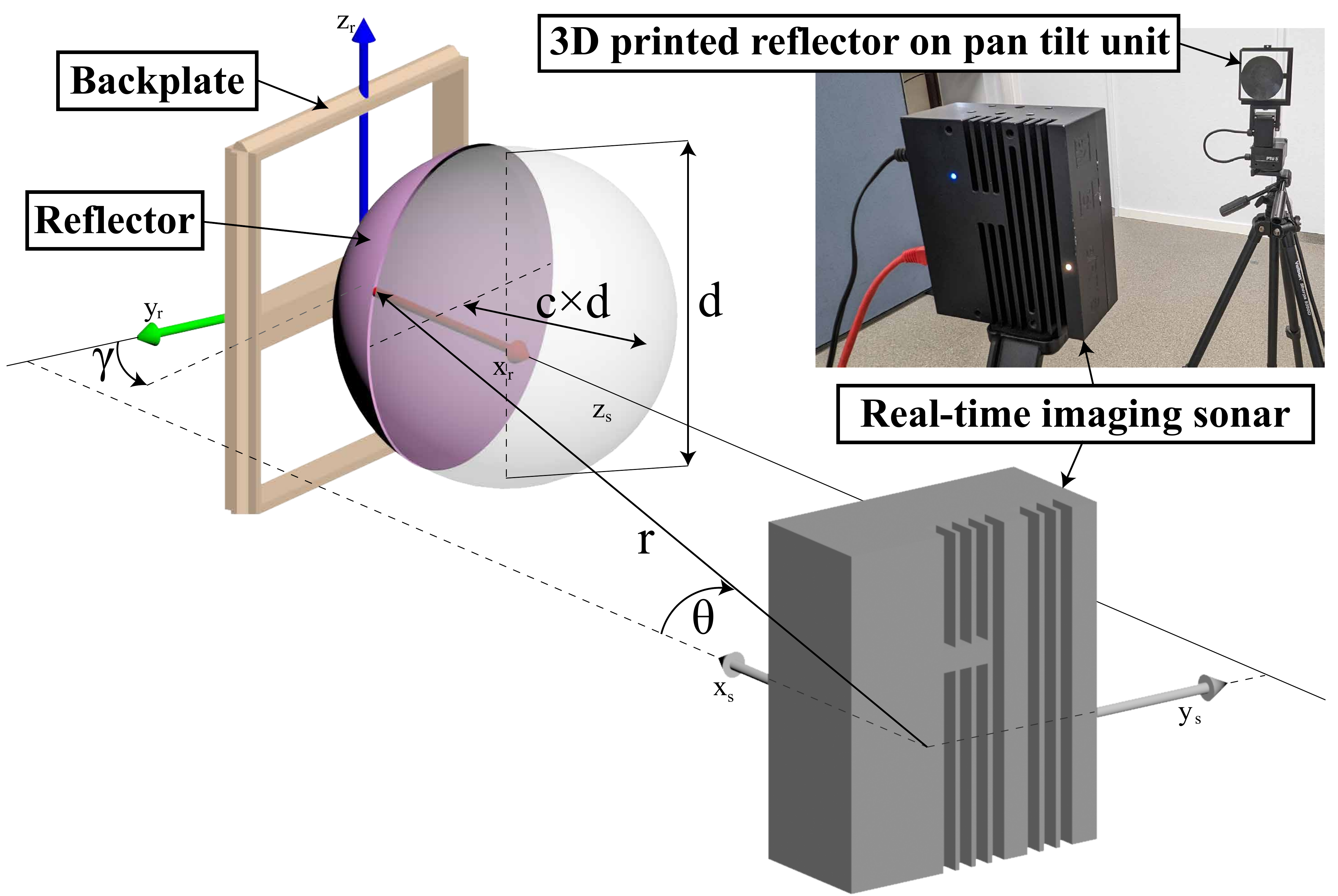}
    \caption{Drawing of a reflector and its dish-shape extracted from a sphere with radius $d$ cut off by a percentage factor $c$. Furthermore, the rotation $\gamma$ is shown. The eRTIS sensor is always assumed to be parallel to the origin of the reflector landmark ($\gamma=0^{\circ}$). The azimuth angle $\theta$ and range $r$ that the sensor observes the reflection at after beamforming is also illustrated. The top right photograph shows the real measurement setup with the pan-tilt device that controls the azimuth and elevation angles of the 3D-printed landmark.}
    \label{fig:sphere}
\end{figure}

\section{Landmark Detection \& Classification} \label{sec:detectionclassification}

\subsection{In-Air 3D Sonar \& Semantic Landmarks}\label{subsec:sonarlandmarks}
The in-air 3D sonar is the Embedded Real-Time Imaging Sonar (eRTIS)\cite{kerstens_ertis_2019}—a device with a 32-element microphone array and a single transducer for broadband signal emission. The MEMS microphones are in a known, irregularly scattered layout. It is capable of scanning the frontal hemisphere of the sensor. In this research it recorded for \SI{36.4}{\ms} at a measurement frequency of \SI{450}{\kilo\hertz}. The emitter casts a broadband FM-sweep of \SI{2.5}{\ms} between \SI{25}{\kilo\hertz} and \SI{80}{\kilo\hertz} for an increased spatial resolution. 
\\The semantic landmarks are based on the original research by Simon et al. \cite{simon_bioinspired_2020}. In this research, we have 3D printed them in ten different radii variations, all with tapered edges. The radius of the spheres were set between \SI{10}{\mm} and \SI{55}{\mm} with increments of \SI{5}{\mm}. The original sphere shape with diameter $d$ is cut at a certain percentage factor $c$. In this paper, this is always 66\%. The landmarks were printed on a backplate for easier mounting. The reflector is illustrated in Figure \ref{fig:sphere}. 

\begin{figure}
    \centering
    \includegraphics[width=0.8\linewidth]{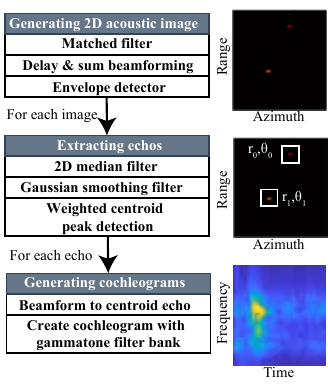}
    \caption{The pre-processing steps for creating the datasets from the microphone-array signals.}
    \label{fig:cochleogram_dataflow}
\end{figure}

\subsection{Data Pre-Processing}\label{subsec:datapreprocessing}
When the eRTIS sensor performs measurements, the goal is first to find potential echoes that can be used as input for the neural network. To this end, a set of standard signal processing steps is taken by first using a matched filter on all 32 microphone signals for improving the SNR. Then, we use conventional delay-and-sum beamforming to steer the array into directions of interest across the sensor's entire frontal hemisphere and find the echoes' location. An envelope detector of \SI{1}{\kilo\hertz} is used to clean up the spatial acoustic image. \\ For further narrowing down the possible echoes within the acoustic image, an additional combination of a median and Gaussian smoothening filter is used. \\ Finally, for fast detection of the echoes, we use a weighted centroid peak detection on the acoustic image to locate potential echo locations and their range and azimuth coordinate pair $(r_i,\theta_i)$. For each of these located echoes, the microphone array is once again steered towards the direction of this echo to receive a single signal, which is windowed only to contain the reflected echo. Next, as the input to our neural network, we propose the use of a cochleogram modeled after the inner ear of a biological species such as an echolocating bat \cite{valero_gammatone_2012, patterson_r_d_efficient_1987}. It uses a non-uniform spectral resolution by associating wider bandwidths with higher frequencies to extract more spectral information \cite{gao_cochleagram-based_2014}. We use a gammatone filter bank to decompose a signal by passing it through a bank of 40 gammatone filters equally spaced on the equivalent rectangular bandwidth (ERB) scale. between \SI{20}{\kilo\hertz} and \SI{110}{\kilo\hertz}. Figure \ref{fig:cochleogram_dataflow} schematically shows the entire processing pipeline.

\begin{figure*}
    \centering
    \includegraphics[width=1\linewidth]{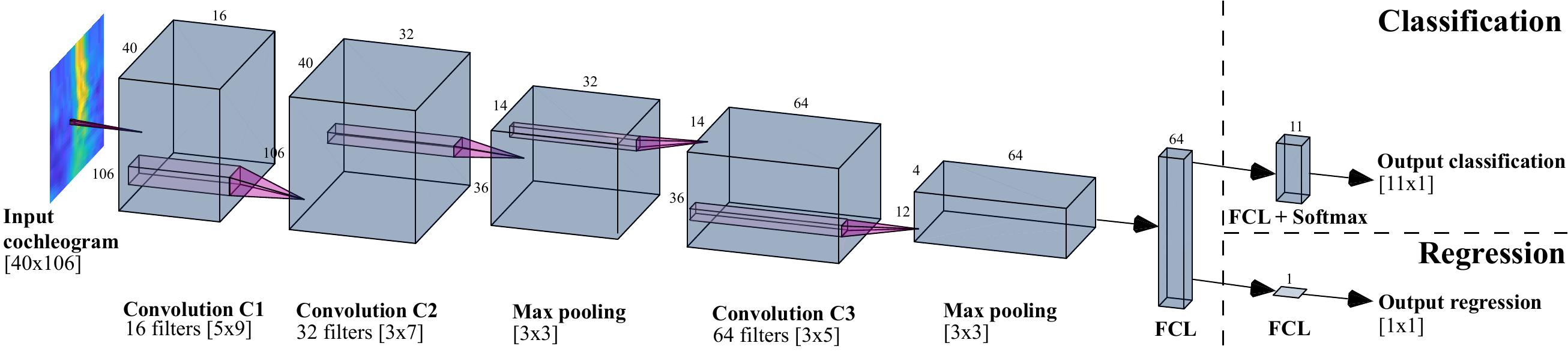}
    \caption{The Neural-network architecture from the input cochleogram for identifying the landmark reflectors and estimating its orientation using a classification and regression network, respectively. These two network outputs share a common architecture of three convolutional layers as well as two Fully Connected Layers (FCL). Batch normalization and rectifier layers were also used but were omitted from the diagram for readability.}
    \label{fig:network_architecture}
\end{figure*}

\subsection{Network Architecture}\label{subsec:detectionclassification}
The main two goals of the neural network are to detect and classify the ten different landmark reflectors and, if there is a reflector present, to estimate its orientation $\gamma$. This is illustrated in Figure \ref{fig:sphere} as well. We used a convolutional neural network of several layers with each echo's cochleogram as input in the form of a 40 by 106-pixel image. After the convolutional layers, fully connected layers are used to either get a classification output (with a softmax function) or the regression output with the orientation of the potential reflector. The entire network architecture is detailed and shown in Figure \ref{fig:network_architecture}. A learning rate of 0.001 was used with the ADAM optimization algorithm. A maximum of 100 epochs was set, but validation patience was set to 20 epochs. For the classification network, class weights were used based on their frequency in the training data.

\begin{figure}
    \centering
    \includegraphics[width=0.98\linewidth]{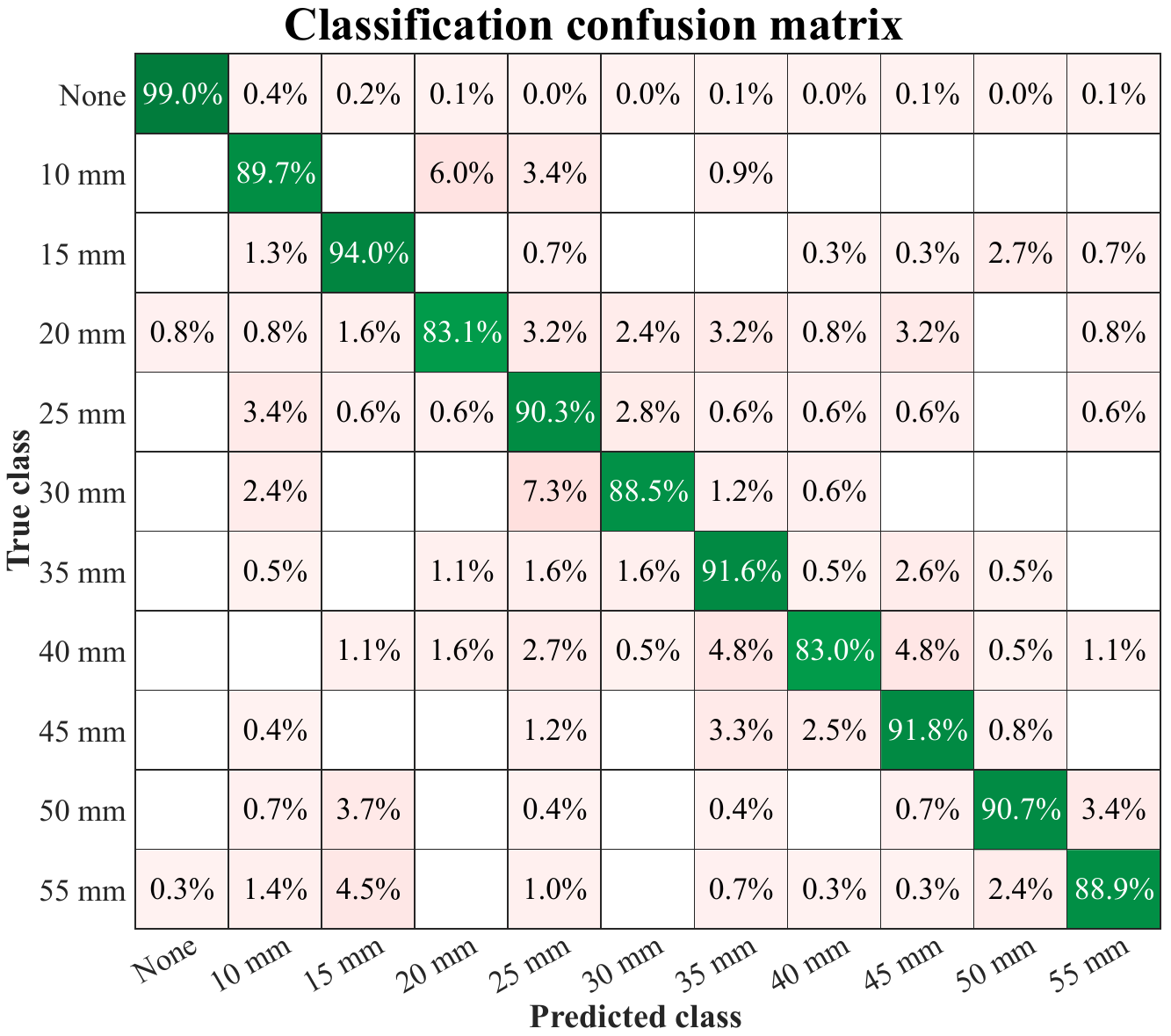}
    \caption{The landmark classification results for ten different landmark sizes and no landmark in the echo in the form of a confusion matrix.}
    \label{fig:single_classification_confusion}
\end{figure}

\begin{figure}
    \centering
    \includegraphics[width=0.98\linewidth]{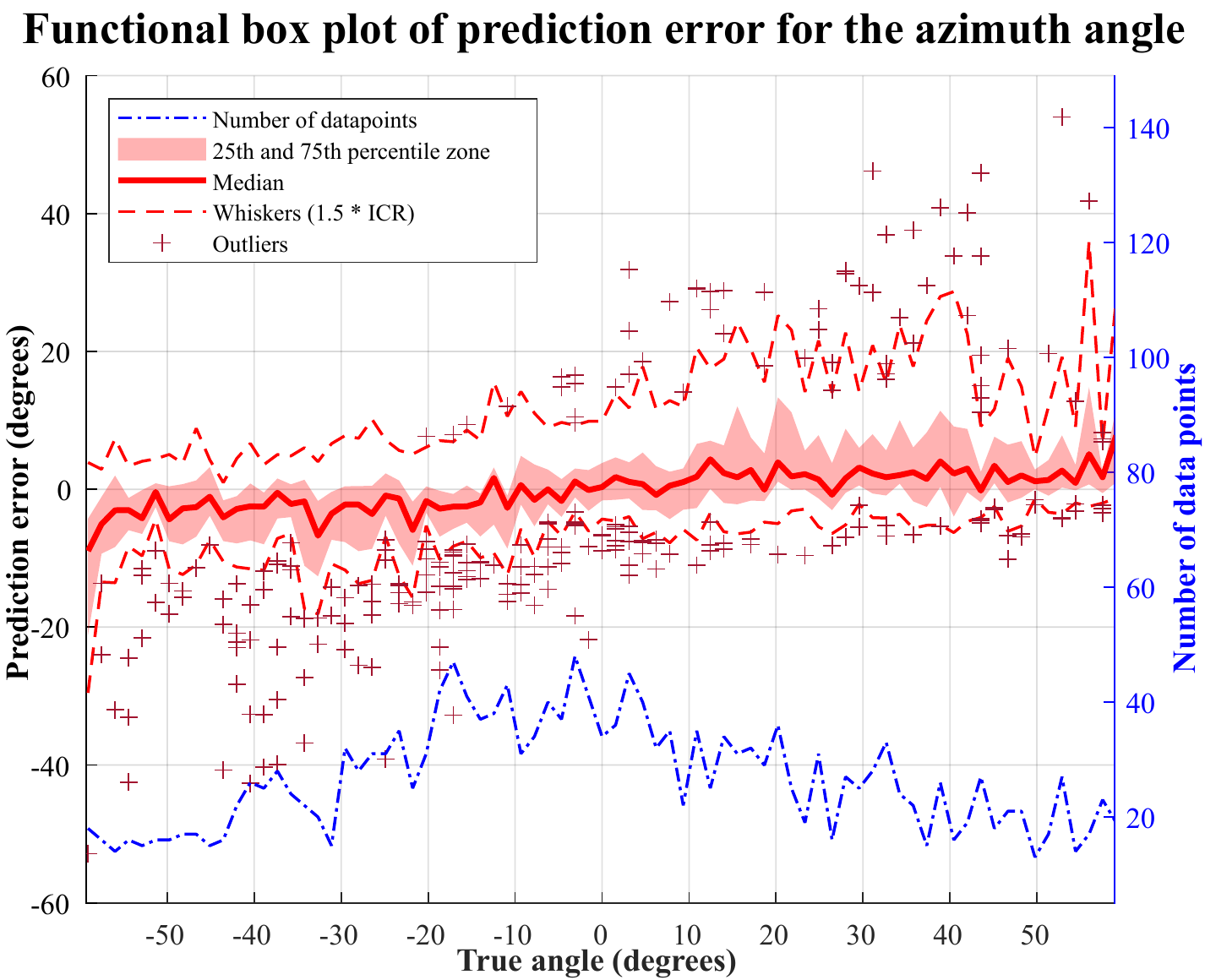}
    \caption{The functional box plot results of the regression for determining the azimuth angle of the landmark.}
    \label{fig:multi_regression_azimuth_boxplot}
\end{figure}

\section{Experimental Results}\label{sec:results}
Experiments were performed in a cluttered indoor environment. Measurements were made with the landmark reflector on a pan-tilt unit mounted on a tripod. The pan-tilt unit varied the horizontal orientation of the landmark between \SI{\pm60}{\degree}. The eRTIS sensor was placed within the same environment in various positions to a maximum offset of \SI{3}{\m}. Note that the sensor was always on the same axis facing forward towards the tripod of the landmark reflector. Figure \ref{fig:sphere} illustrates this configuration. Next to the measurements with a landmark reflector, empty measurements were also made by placing the eRTIS sensor on the pan-tilt unit in various locations in the same room and letting it randomly make measurements in various directions. After data pre-preprocessing, 10.293 cochleograms were made with one of the ten reflector types and 46.901 cochleograms were made without any reflector in the room. The data was split into 64\% training, 16\% validation, and 20\% testing. The classification network reached an accuracy of 97.3\% on the test dataset. The confusion matrix of this result is shown in Figure \ref{fig:single_classification_confusion}. These results also show a very high accuracy for the detection with 99\% for classifying the \textit{empty} class. For the regression network to predict the azimuth orientation angle, the result gave an RMSE of \SI{9.1528}{\degree}. Figure \ref{fig:multi_regression_azimuth_boxplot} shows a functional box plot for the entire azimuth range, indicating an apparent degradation towards the outer angles.

\section{Conclusions \& Future Work}\label{sec:conclusions}
The experimental results show that the proposed neural network architecture with the cochleogram representation of ultrasonic echoes can accurately detect and classify the landmark reflectors with very closely clustered variations of these landmarks with a maximum difference in diameter of only \SI{5}{\mm}. Furthermore, its orientation angle could be sufficiently accurately predicted for applications such as SLAM. By also being able to estimate the orientation of these landmarks after detecting and classifying them, their capability for minimizing the error in, for example, SLAM or autonomous navigation algorithms increases. However, these results also show the potential for increasing the encoded information within these acoustic landmarks by creating a priori known patterns with them.
\clearpage
\bibliographystyle{IEEEtran}
\bibliography{main.bib}

\end{document}